\pdfoutput=1
\documentclass[11pt]{article}
\usepackage[preprint]{acl}
\usepackage{times}
\usepackage{latexsym}
\usepackage{booktabs}
\usepackage{amsmath}
\usepackage{amssymb}
\usepackage{graphicx}
\usepackage{tikz}
\usetikzlibrary{arrows.meta,positioning,fit}
\usepackage{microtype}
\usepackage{url}

\title{Staged Linguistic Seeding: Grounded Query Expansion for\\
Verified-Unit QA in AI Contact Centers}

\author{
  Hyeonseop Yoon\thanks{Equal contribution.} \\
  Independent Researcher \\
  Seoul, South Korea \\
  \texttt{hyeonseopy@acm.org}
  \And
  Jeong-Eun Park\footnotemark[1] \\
  Independent Researcher \\
  Seoul, South Korea \\
  \texttt{20181053@dongduk.ac.kr}
}

\begin{document}
\maketitle

\begin{abstract}
Customer-service QA in an AI contact center (AICC) runs under deployment constraints that benchmark QA
misses: tight voice-hotline latency and a high cost for unsupported or wrong automatic answers. We
deploy a system that answers only from a \emph{closed set} of \emph{verified QA units}: it returns a
retrieved unit verbatim, or routes to clarify/abstain/handoff. The index is enriched \emph{offline} by
\emph{staged linguistic seeding} (SLS): a human authors a per-unit world-grounded slot recipe,
\texttt{gpt-4.1-mini} renders it into variants, and a light human gate filters them, one methodology
reused across both domains, so inference stays a single retrieval pass with no query-time generation. On held-out query
variants (in-distribution recipe reformulations) from two industrial domains, SLS is the dominant retrieval lever, lifting hybrid \texttt{R@1} to
\textbf{0.881}/\textbf{0.930} ($+0.27$/$+0.34$); the gain holds across all five retrievers we test. The
win is the design, not the volume: at the \emph{same} \texttt{gpt-4.1-mini}, SLS beats a budget-matched
doc2query by $+0.20$/$+0.32$, and a
cross-provenance control provides additional evidence of transfer across generated-query distributions; the embedding choice
is secondary. Verified-unit answering also
removes free-form generation's unsupported-content surface ($7$--$13\%$ vs.\ ${\approx}0\%$). We report
this as an honest application study, including negative results.
\end{abstract}

\section{Introduction}

Large language models and retrieval-augmented generation (RAG) dominate standard QA benchmarks
\citep{lewis2020retrieval}. But a production AI contact center (AICC), long automated with virtual
agents \citep{gilbert2005ivagent} and increasingly built on LLM agents \citep{choubey2025serviceagents},
imposes constraints those benchmarks never test, especially on a voice hotline. The cost of errors is \emph{asymmetric}. A confident but
unsupported answer to an already-frustrated caller is far more damaging than a clarifying question or a
handoff to a human agent \citep{banerjee2023customersupport,yu2020financialchatbot}. Latency makes this
harder. On a live voice line the budget is tight, so query-time generation (a per-call LLM round trip
of roughly a second) is impractical. Our deployment goal is concrete: \emph{relieve} human agents, not
replace them, keeping a safe escape to a person without escalating every ambiguous case.

RAG is the default response, because it grounds answers in approved references. But grounding the
retriever is not enough: as long as a free-form generator is licensed to emit text, it can over-extend
even correct evidence and assert unsupported claims \citep{niu2024ragtruth}. We measure this directly. Over the
\emph{same} retrieved evidence, free-form RAG answers are judged unsupported in $7$--$13\%$ of cases
(\texttt{gpt-4o} judge). This residual persists even on correct evidence, and we did not find it tunable to zero; removing
generation from the serving path eliminates it by construction.

We built and deployed a \emph{verified-unit} QA system that excludes generation entirely. It
answers only from a closed set of human-verified QA units (each a representative question, a verified
answer, and a unit identifier), returning a retrieved unit's answer verbatim, or routing to clarify,
abstain, or handoff. Nothing is generated at serve time, so unsupported answers are zero \emph{by
construction}. But removing the generator opens a new gap. Callers phrase one need in countless ways,
while each unit ships with a single representative question. Without a query-time paraphraser, the full
linguistic distance between \emph{user-query diversity} and the curated answer set falls on retrieval.
Our core contribution, \emph{Staged Linguistic Seeding} (SLS), closes that gap offline. A human authors
a per-unit, world-grounded slot recipe; \texttt{gpt-4.1-mini} renders it into candidate variants; a
light human gate keeps or drops each one. The recipe injects domain knowledge, the LLM supplies surface
variety, and the seeding axes and prompt are reused unchanged across both domains. Everything runs
offline, so inference stays a single retrieval pass with zero added serving cost.

This is an application paper. We built a verified pipeline that addresses
two production constraints (unpredictable generation and tight latency), and we give empirical evidence
for \emph{why} it works on two real anonymized enterprise domains, with the core gain (and a closed-set
leakage pitfall) replicating on a public benchmark. Our contributions, ordered by impact, are:
\begin{enumerate}
  \item \textbf{A generation-free verified-unit pipeline in production use.} We built a closed-set
    QA system, in production use at an AICC (${\approx}74$k requests, ${\approx}89\%$ contained), over two anonymized domains (\textsc{Auto}, $90$ units; \textsc{Elec},
    $229$ units; $319$ verified units and $7{,}947$ query variants in total), evaluated with a
    unit-attribution metric (\texttt{gold\_unit\_id} at \texttt{R@1}) on leakage-controlled held-out
    query variants \citep{rashkin2021attributable}; a faithfulness study quantifies the unsupported-content
    surface it removes ($\approx 0\%$ vs.\ $7$--$13\%$ for free-form RAG on identical evidence).
  \item \textbf{SLS is the dominant retrieval lever, and we isolate \emph{why}.} Offline expansion
    lifts hybrid (BM25+BGE-M3) \texttt{R@1} to $0.881$ (\textsc{Elec}) and $0.930$ (\textsc{Auto}),
    $+0.27$/$+0.34$ over a single representative question, and the gain holds across all five retrievers
    \citep{robertson2009probabilistic,chen2024bge}. A matched-budget control isolates the
    cause: with the \emph{same} \texttt{gpt-4.1-mini}, SLS beats a budget-matched doc2query by
    $+0.20$/$+0.32$ and every automatic baseline (doc2query, HyDE, query2doc) by at least
    $+0.20$/$+0.29$ \citep{nogueira2019document,gao2022precise,wang2023query2doc}. The lever is the seeding
    \emph{design}, not the model, the augmentation family, or the volume of text. Serving stays a single
    pass ($\sim$14\,ms hybrid vs.\ $\sim$1\,s for query-time generation).
  \item \textbf{The mechanism is consistent with broader phrasing coverage.} A cross-provenance
    control shows asymmetric transfer: the SLS index is $0.146$ below doc2query on doc2query-held-out
    queries, whereas the doc2query index is $0.399$ below SLS on SLS-held-out queries. Because the
    resulting index sizes differ, we treat this control as suggestive rather than volume-controlled. A symbol-grounding analysis sharpens
    this: $49\%$ of held-out \textsc{Auto} queries share \emph{zero} content words with the canonical
    question, and SLS lifts that surface-disjoint half by $+0.59$, the gain decaying monotonically to
    $+0.01$ as phrasings approach full overlap \citep{harnad1990}.
  \item \textbf{External validity and honest negatives.} Both protocol-level findings (the expansion
    gain \emph{and} a closed-set self-retrieval leakage pitfall) replicate, in sign and more modestly, on the public Quora Question
    Pairs benchmark \citep{quora_qqp,lewis2021testtrain}, beyond our in-house logs. The embedding choice
    is secondary: neither a larger Qwen3 \citep{zhang2025qwen3embedding} nor a Korean-specialized
    fine-tune beats off-the-shelf BGE-M3. Finally, a class-balanced router on non-leaky
    retrieval-confidence features recovers the rare clarify/abstain/handoff actions (action macro-F1
    $0.36\!\to\!0.51$), but stays bounded by heuristic labels.
\end{enumerate}

\section{Related Work}

\paragraph{Closed-set / FAQ retrieval.}
Answering from a fixed set of verified question--answer units is the FAQ-retrieval setting
\citep{sakata2019faq,mass2020faq}; \citet{mass2020faq} expand FAQ entries with generated questions. We
add a unit-ID attribution metric, leakage-controlled held-out-query evaluation, and an explicit
non-answer action space on real two-domain enterprise logs. Restricting the emitted answer to a closed,
approved set (as in constrained generation over a fixed candidate set \citep{decao2021autoregressive}
and verified-quote answering with abstention \citep{menick2022gophercite}) makes non-fabrication a
property of the \emph{answer space}, not the generator.

\paragraph{Grounded retrieval and attribution.}
Retrieval alone does not prevent unsupported claims
\citep{lewis2020retrieval,rashkin2021attributable,gao2023alce,es2023ragas,saadfalcon2023ares}, and
RAGTruth \citep{niu2024ragtruth} documents hallucination even on correct context. We instantiate
attribution-style evaluation \citep{rashkin2021attributable} as unit-ID recovery and restrict the
answer surface to verified units.

\paragraph{Query expansion and offline augmentation.}
doc2query, query2doc, and HyDE show that LLM-generated text can improve retrieval, though unevenly and not on every retriever
\citep{nogueira2019document,wang2023query2doc,gao2022precise,jagerman2023queryexpansion,riabi2021synthetic}.
Following EnrichIndex \citep{chen2025enrichindex}, recent work \citep{ma2025drama,su2025parametricrag} moves
this enrichment offline, atop a long classical query-expansion lineage
\citep{rocchio1971relevance,lv2010positional,carpineto2012survey,azad2017queryexpansion,bhogal2007ontology,massai2022semanticrelations}.
Our expansion (SLS, \S\ref{par:aug}) is offline doc2query-style enrichment, but it is \emph{seeded} by a
world-grounded slot recipe under a reusable methodology rather than by single-shot model output, then gated by a light human
check; such linguistically motivated, in-distribution augmentation can pay off, particularly in lower-resource settings
\citep{groshan2025linguistic}, and no generation happens at serving time.

\paragraph{Retriever families.}
The main retriever families (BM25, SPLADE, BGE-M3, and the dense Qwen3
\citep{robertson2009probabilistic,formal2021splade,chen2024bge,karpukhin2020dense,zhang2025qwen3embedding}) span
the lexical--semantic spectrum; we take one representative from each and compare them under a single
verified-unit protocol.

\paragraph{Abstention and routing.}
Treating refusal and handoff as first-class actions is the premise of selective classification, refusal
tuning, escalation, and cost-sensitive decision theory
\citep{geifman2017selective,zhang2024rtuning,kirichenko2025abstentionbench,yu2020financialchatbot,banerjee2023customersupport,elkan2001costsensitive},
whereas LLM routing and cascades instead target \emph{models} \citep{ong2025routellm,chen2023frugalgpt}. Our
actions route the query out of the automatic system under asymmetric costs.

\section{Problem Setup}

Given a human query $x$, a verified QA-unit index $T=\{t_i\}$, and optional metadata, the system
chooses an action $a \in \{\textsc{answer}, \textsc{clarify}, \textsc{abstain}, \textsc{handoff}\}$.
When $a=\textsc{answer}$, it returns the answer of a selected unit $t_i$, so that the retrieval target is
the unit identifier rather than a generated string. Each unit is stored as
$t_i=(\mathrm{unit\_id}, q_i, y_i)$, where $q_i$ is a representative question and $y_i$ a
verified answer: the stored record is the implementation object, while the verified QA unit is the
conceptual grounding boundary.

\paragraph{Why verified units bound the failure surface.}
Let $i^\star$ be the gold unit for query $x$, let $\hat{\imath}=\arg\max_i \operatorname{score}(x,t_i)$
be the retrieved top-1 unit, and let $\rho=\Pr[\hat{\imath}\neq i^\star]=1-\texttt{R@1}$ be the
retrieval-error rate. Call an answer \emph{unsupported}, $U(a)=1$, if it asserts content not present
in the retrieved evidence $E(x)$. A free-form answer $a=g(x,E(x))$ from a generator $g$ can be
unsupported even when the evidence is correct, at a rate $\eta=\Pr[U(a)\mid \hat{\imath}=i^\star]>0$
(documented by RAGTruth \citep{niu2024ragtruth}; we measure $\eta\!\approx\!0.12$ in
\S\ref{sec:faithfulness}). Its unsupported rate is therefore
\begin{equation}
\Pr\nolimits_{\text{free}}[U]=(1-\rho)\,\eta+\rho\,\eta' \;>\;0,
\label{eq:free}
\end{equation}
where $\eta'\!>\!0$ is the unsupported rate under wrong evidence. Verified-unit answering instead
returns $a=y_{\hat{\imath}}$, the human-verified text of a retrieved unit, so the answer is always an
element of the evidence and cannot introduce new claims:
\begin{equation}
\begin{aligned}
a\in\{y_i : t_i\in E(x)\}\;\Longrightarrow\;\Pr\nolimits_{\text{ours}}[U]=0,\\
\Pr\nolimits_{\text{ours}}[\text{error}]=\rho.
\end{aligned}
\label{eq:ours}
\end{equation}
Comparing \eqref{eq:free} and \eqref{eq:ours}, the verified-unit design removes the open-ended generation-error term $\eta$ and leaves a single
\emph{measurable} retrieval-error term $\rho=1-\texttt{R@1}$, on whose low-confidence part the router
(\S\ref{sec:routing}) declines to answer. This is ``non-fabrication by construction'': we eliminate not
\emph{wrong} answers (a wrong unit can be retrieved) but \emph{unsupported} ones.

\section{Data Construction}

The project uses real enterprise QA logs from two anonymized deployments, \textsc{Auto} (automotive)
and \textsc{Elec} (consumer-electronics). These are not benchmark pairs: rows include short turns,
incomplete intent, domain shorthand, greetings, URL/image prompts, and many queries whose correct
action is clarify, abstain, or handoff. From these real logs we derive two artifacts.

\paragraph{(a) Verified units with query-expansion variants (retrieval study).}
The core asset is a curated FAQ in which each verified unit carries a verified answer together with query
variants produced by SLS (\S\ref{par:aug}) and seeded from the representative question and the phrasings
real users and operators recorded in the logs. Because the variants are an \emph{output} of the protocol
rather than hand-authored strings, their per-unit count in the full pre-split pool ($5$--$113$; indexed-only counts of $3$--$79$ in App.~\ref{app:data}) tracks the question's linguistic variability. This yields
\textbf{319 verified units} (229 \textsc{Elec}, 90 \textsc{Auto}) and \textbf{7{,}947 query variants}, whose held-out
portion forms our leakage-free test set of real-log-seeded variants (held-out protocol variants, not raw logged utterances). The protocol was applied to \emph{both} domains; the explicit
authoring recipe is retained for \textsc{Auto} (App.~\ref{app:example}), while for \textsc{Elec}, owing
to data-retention constraints, only the deployed variants remain.

\paragraph{(b) Operational log with action labels (routing study).}
For routing we use a larger operational set of $6{,}000$ clustered human queries with heuristic action
labels (Table~\ref{tab:data}), built by cleaning and deduplicating raw logs, clustering similar
questions, and assigning answerability/routing labels. Retrieval is therefore measured on the held-out variants from
(a), and routing on the labeled log (b).

\paragraph{Data governance.}
Because the logs are enterprise data, we report only aggregate statistics, anonymized examples, and
unit IDs; raw user text remains access-controlled. Privacy is part of the system boundary, not an
appendix detail.

\section{System}

\begin{figure*}[t]
\centering
\resizebox{0.82\textwidth}{!}{%
\begin{tikzpicture}[
  font=\footnotesize,
  b/.style={draw,align=center,inner sep=3pt,minimum height=7mm},
  hum/.style={b,fill=black!8},
  st/.style={b,fill=black!4},
  ref/.style={draw,circle,inner sep=1.5pt,minimum size=5mm,fill=black!10},
  sf/.style={draw,rounded corners=1pt,inner sep=2pt,font=\scriptsize,minimum height=4.5mm},
  ar/.style={-{Latex[length=1.6mm]}},
  node distance=3mm and 5mm,
]
\node[font=\scriptsize\bfseries] (gl) at (0,0) {Symbol grounding (form vs.\ meaning)};
\node[sf,below left=5mm and 1mm of gl] (f1){``set GPS to get there''};
\node[sf,below=5mm of gl] (f2){``address''};
\node[sf,below right=5mm and 1mm of gl] (f3){``where is it''};
\node[ref,below=6mm of f2] (r){};
\node[right=2.5mm of r,font=\scriptsize,align=left] {one \emph{referent}\\(meaning)};
\foreach \f in {f1,f2,f3}{\draw[ar](\f)--(r);\draw[ar](r)--(\f);}
\node[font=\scriptsize\itshape,align=center,below=2mm of r,text width=58mm]
 {operator grounds diverse \emph{surface forms} in one referent (world knowledge);\\
  a thesaurus stays surface$\leftrightarrow$surface (dictionary-go-round) and misses surface-disjoint forms.};
\node[draw,dashed,fit=(gl)(f1)(f3)(r),inner sep=4pt] (insetbox){};
\node[b,right=46mm of gl,text width=20mm] (q0){representative\\question $Q_u$\\(+ answer $A_u$)};
\node[hum,below=5mm of q0,text width=40mm] (rec){\textbf{SLS recipe} (human):\\ factorize $Q_u$ into grounded slots\\ $S_1, S_2, \dots, S_k$\\ {\scriptsize(each slot $=$ one referent, many forms; App.~\ref{app:example})}};
\node[b,below=5mm of rec,text width=40mm] (vars){\texttt{gpt-4.1-mini} recombines $+$ gate\\ $\Rightarrow$ variants $q^u_1,\dots,q^u_n$};
\draw[ar](q0)--(rec);\draw[ar](rec)--(vars);
\draw[ar,dashed] (insetbox.east|-rec) -- node[midway,above,font=\scriptsize]{grounds}(rec.west);
\node[font=\scriptsize\itshape,above=0.4mm of q0]{OFFLINE: build index};
\node[st,right=10mm of rec,text width=21mm] (idx){\textbf{verified-unit index}\\$\{q^u_1\dots q^u_n\}\!\mapsto\! A_u$};
\draw[ar](vars.east) to[out=0,in=-110] (idx.south);
\node[b,below=7mm of vars,text width=18mm] (uq){user query $q$};
\node[b,right=of uq,text width=28mm] (retr){retrieve: match $q$ vs variants $\{q^{*}_i\}$ (BM25/dense/hybrid) $\to\hat u$};
\node[b,right=of retr,text width=15mm] (rt){cost-sens.\\router};
\node[st,right=13mm of rt,text width=20mm] (ansv){$A_{\hat u}$ \textbf{verbatim}\\{\scriptsize(non-fabrication)}};
\node[b,below=5mm of ansv,text width=20mm] (esc){clarify/abstain/\\handoff};
\draw[ar](uq)--(retr);\draw[ar](retr)--(rt);
\draw[ar](rt.east)-- node[midway,above,font=\scriptsize]{conf.}(ansv.west);
\draw[ar](rt.south)|- node[pos=.75,above,font=\scriptsize]{weak}(esc.west);
\draw[ar](idx) to[out=0,in=90] node[right,font=\scriptsize,pos=.5]{index}(retr.north);
\node[font=\scriptsize\itshape,left=0.5mm of uq,rotate=90,anchor=south]{ONLINE: serve};
\end{tikzpicture}}
\caption{System overview. \textbf{Offline}, a human operator factorizes each verified unit's
representative question $Q_u$ into world-grounded slots---\emph{symbol grounding}: diverse surface
forms tied to one referent, a link a thesaurus cannot make (it stays surface$\leftrightarrow$surface)---and
\texttt{gpt-4.1-mini} recombines the slots into the indexed variants $q^u_1\dots q^u_n$. \textbf{Online},
a user query $q$ is matched against the \emph{variants} (not the canonical question), and a
cost-sensitive router returns the verified answer verbatim or escalates. The grounded variants bridge
the gap between curated canonical questions and the surface-disjoint ways real users phrase the same
intent (\S\ref{sec:discussion}).}
\label{fig:overview}
\end{figure*}
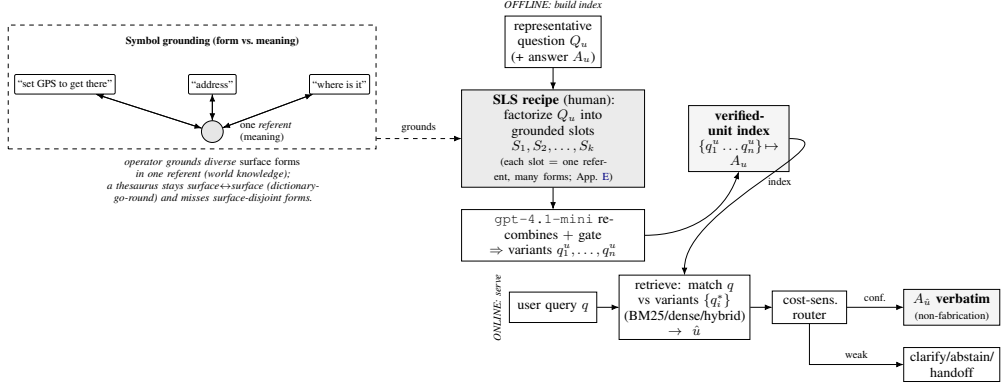

\paragraph{Verified-unit retrieval.}
Figure~\ref{fig:overview} overviews the system: offline index construction by staged linguistic seeding
(a human operator grounds each unit's representative question in world-grounded slots, which
\texttt{gpt-4.1-mini} recombines into the indexed variants) and online verified-unit retrieval with
cost-sensitive routing. We compare one retriever per family over the same verified QA units, spanning
the lexical--semantic spectrum: BM25 \citep{robertson2009probabilistic} as the sparse baseline; the
dense-only Qwen3 embedding \citep{zhang2025qwen3embedding}; and BGE-M3 \citep{chen2024bge} in
between, a multi-functional encoder (dense, sparse, and multi-vector modes) of which we use the dense
mode. The hybrid linearly combines per-query min--max-normalized BM25 and BGE-M3 scores at equal
weight. For the dense component we use the official BGE-M3 checkpoint, more robust on this corpus than
the available domain-adapted alternative (\S\ref{sec:results}).

\paragraph{Offline query expansion: staged linguistic seeding (SLS).}
\label{par:aug}
For each verified unit the index stores either the representative question (RAW) or that question plus
expansion variants (AUG). The variants are produced offline by \emph{staged linguistic seeding}
(Figure~\ref{fig:overview}), which separates what is reused from what is authored per unit. Reused
across units and both domains are (i)~\textbf{seeding axes}, a fixed specification of the linguistic
and world-grounded forms to cover (synonyms, homonyms, syllable and lemma/stem forms, and the
\emph{eojeol}, the Korean space-delimited token)---together with a single fixed \textbf{prompt} conditioning
\texttt{gpt-4.1-mini} (the \emph{same} model the automatic baselines use) on those axes. Per unit, a
human then authors a \textbf{world-grounded slot recipe} that factorizes the representative question
into grounded slots along these axes; (ii)~\texttt{gpt-4.1-mini} \textbf{renders} it into candidate strings, which
(iii)~a \textbf{light human gate} keeps or drops against a fixed rubric. The durable asset is the
recipe (from which the indexed variants are regenerable on demand), not the $7{,}947$ output strings.
SLS is thus doc2query-style \citep{nogueira2019document} in structure; the lever is the human,
world-grounded seeding \emph{distribution} (\S\ref{sec:results}), and inference stays a single retrieval pass.

\paragraph{Cost-sensitive routing (the policy we test).}
Retrieval alone is insufficient when a query is incomplete, unsupported, or risky, so we add a
cost-sensitive decision over \emph{non-leaky} retrieval-confidence features (per-family top/top-2
scores, margins, cross-family agreement), selecting the expected-cost-minimizing action under a cost
matrix where a wrong answer costs far more than a clarification (a threshold gate is the one-scalar
special case). We estimate it with a class-balanced gradient-boosted classifier and report it as a
bounded secondary study (\S\ref{sec:routing}, App.~\ref{app:routing}).

\section{Experiments}
\label{sec:results}

\paragraph{Metrics and protocol.}
The primary metric is unit recall at rank 1 (\texttt{R@1}) against each variant's gold source unit
(\texttt{gold\_unit\_id}); we also report \texttt{R@3} and \texttt{R@15}. For the retrieval study, for each verified unit we randomly
hold out $30\%$ of its query variants as \emph{test} queries that are never indexed,
and index the remaining variants (AUG) or only the single representative question (RAW). Because the
test phrasing is never in the index, the protocol is leakage-free at the string level. This yields
$1{,}468$ held-out \textsc{Elec} and $915$ held-out \textsc{Auto} test queries over the $319$ units. Because held-out and
indexed variants are drawn from the same per-unit pool, the test is \emph{in-distribution} \citep{lewis2021testtrain}: it measures
generalization to unseen phrasings of each unit's recipe, not transfer to an independent query stream
(on \textsc{Auto}, $49\%$ of held-out queries share no content word with the representative question). A
cross-provenance control provides suggestive evidence that the gains are not solely due to this provenance match.
A fully independent organic test is unavailable by construction: raw call transcripts are not retained,
and operational logs carry routing-action labels rather than unit-level gold, since most organic queries
do not map to a single verified unit (Table~\ref{tab:data}, $59\%$ non-answer).

\paragraph{A leakage check.}
Querying a unit with its own indexed representative question \emph{trivially} inflates \texttt{R@1} (up
to $\approx 8\times$ for BM25); our held-out-variant protocol avoids this by construction; no test
phrasing is ever indexed (audit in App.~\ref{app:leakage}).

\paragraph{Staged linguistic seeding is the dominant lever.}
Table~\ref{tab:main} reports \texttt{R@1} on the held-out query variants (App.~\ref{app:interp} traces these back to the per-domain data statistics). Across both domains and all
retrievers, offline SLS expansion is the dominant gain: the hybrid rises from $0.609\!\to\!0.881$
on \textsc{Elec} ($+0.272$) and $0.588\!\to\!0.930$ on \textsc{Auto} ($+0.342$); BM25 gains most
($+0.351$/$+0.426$), and the gain holds across the full lexical--semantic spectrum (SPLADE
$+0.219$/$+0.253$, Qwen3 $+0.135$/$+0.168$, dense $+0.123$/$+0.202$). The best
configuration is hybrid${+}$expansion (\textbf{0.881}/\textbf{0.930}), with \texttt{R@3}
$0.973$/$0.981$. Every gain is significant by a paired bootstrap (10k resamples, $n{=}1{,}468$/$915$):
hybrid $+0.272$ (95\% CI $[+0.247,+0.297]$) / $+0.342$ $[+0.309,+0.375]$; BM25 $+0.351$
$[+0.324,+0.378]$ / $+0.426$ $[+0.392,+0.461]$; dense $+0.123$ $[+0.101,+0.144]$ / $+0.202$
$[+0.175,+0.231]$. No interval includes zero. Because queries from one unit are not independent, we
also report a \emph{cluster} (per-unit) bootstrap that resamples whole units rather than queries:
intervals widen, as expected, but every gain remains significant, and the smallest-domain headline
---\textsc{Auto} hybrid $+0.342$---survives even at its $90$ units (cluster $95\%$ CI $[+0.291,+0.390]$).

\begin{table}[t]
\centering
\footnotesize
\setlength{\tabcolsep}{4pt}
\resizebox{\columnwidth}{!}{%
\begin{tabular}{lccccccc}
\toprule
& \multicolumn{3}{c}{Elec} & \multicolumn{3}{c}{Auto} \\
\cmidrule(lr){2-4}\cmidrule(lr){5-7}
Retriever & RAW & {+}AUG & $\Delta$ & RAW & {+}AUG & $\Delta$ \\
\midrule
BM25 (lexical)             & 0.471 & 0.822 & {+}0.351 & 0.480 & 0.906 & {+}0.426 \\
SPLADE-ko (learned sparse) & 0.627 & 0.846 & {+}0.219 & 0.640 & 0.893 & {+}0.253 \\
Dense BGE-M3 (official)    & 0.717 & 0.840 & {+}0.123 & 0.661 & 0.863 & {+}0.202 \\
Dense Qwen3-4B             & 0.690 & 0.825 & {+}0.135 & 0.674 & 0.842 & {+}0.168 \\
Dense (fine-tuned, ours)   & 0.623 & 0.779 & {+}0.156 & 0.591 & 0.814 & {+}0.223 \\
\midrule
Hybrid (BM25{+}BGE-M3)     & 0.609 & \textbf{0.881} & \textbf{{+}0.272} & 0.588 & \textbf{0.930} & \textbf{{+}0.342} \\
\bottomrule
\end{tabular}}
\caption{\texttt{R@1} on held-out query variants (Elec $n{=}1{,}468$, Auto $n{=}915$;
$319$ units), across the lexical--semantic spectrum: lexical BM25, learned-sparse SPLADE, dense
BGE-M3 and Qwen3-4B, and the lexical${+}$semantic hybrid. RAW indexes one representative question per
unit; {+}AUG additionally indexes the unit's deployed SLS variants. Offline SLS
expansion is the dominant lever in \emph{every} family (hybrid $+0.272$/$+0.342$; BM25
$+0.351$/$+0.426$; SPLADE $+0.219$/$+0.253$; Qwen3 $+0.135$/$+0.168$), and the hybrid is best. A larger
Qwen3-4B and our own fine-tuned checkpoint (\texttt{dragonkue/BGE-m3-ko} further fine-tuned on
\textsc{Nemotron-Personas-Korea} persona pairs) do not exceed the official BGE-M3, and the fine-tuned
one trails it throughout, despite starting from a Korean-specialized base (Appendix~\ref{app:ckpt}).
Test phrasings are never indexed, so the setting is leakage-free.}
\label{tab:main}
\end{table}

\paragraph{The embedding choice is secondary to the expansion.}
Down the RAW columns, dense ($0.66$--$0.72$) beats BM25 ($0.47$--$0.48$): the low RAW recall is
expected: the representative question is an expert-prepared canonical form that real phrasings rarely
match lexically (the form--meaning gap, \S\ref{sec:discussion}), and a semantic retriever only partly
bridges it. With expansion, BM25 catches up ($0.82$--$0.91$) and the hybrid is best, so \emph{which}
embedding one picks becomes secondary to the expansion. Across the spectrum the off-the-shelf official
BGE-M3 is strongest: a larger Qwen3-4B does not exceed it (AUG $0.825$/$0.842$ vs.\ $0.840$/$0.863$),
and a Korean-specialized fine-tune (App.~\ref{app:ckpt}) underperforms it in every cell. This is not a training failure but evidence for the protocol: neither a larger
model nor extra generic in-domain data closes the gap that grounded expansion does; the gap is
\emph{coverage of unseen phrasings}, not embedding capacity.

\paragraph{The lever is the seeding design, not the volume: a matched-budget control.}
A natural objection is that \emph{any} generated expansion would help, or that our gain merely buys
more effort per unit. We test both with an offline-vs-online comparison on the same held-out queries
(Table~\ref{tab:bakeoff}): SLS vs.\ offline automatic expansion (doc2query
\citep{nogueira2019document}, EnrichIndex \citep{chen2025enrichindex}) and query-time online
augmentation (HyDE \citep{gao2022precise}, query2doc \citep{wang2023query2doc}), all of them generated by
the \emph{same} \texttt{gpt-4.1-mini} that SLS uses (prompts in App.~\ref{app:baselineprompts}). SLS dominates every automatic method by a wide margin:
hybrid \texttt{R@1} $0.881$/$0.930$ vs.\ the best automatic baseline $0.685$/$0.639$, a
$+0.20$/$+0.29$ gap, at zero query-time generation cost. The decisive evidence is the
\emph{matched-budget} control: when the automatic baseline generates the \emph{same number} of
questions per unit as SLS (``doc2query $N$-matched''; Table~\ref{tab:bakeoff}); i.e.\ the same
generator at the same per-unit budget, it reaches only $0.685$/$0.611$, only $+0.001$/$+0.038$ above the $6$-shot
version and far below SLS (matched gap $+0.196$ $[+0.162,+0.230]$ / $+0.319$ $[+0.240,+0.398]$,
per-unit cluster bootstrap). The gain is therefore \emph{not} bought by indexing more text: with the
generator and budget fixed, the only difference is \emph{how} the variants are seeded: the
world-grounded recipe, not the model, the augmentation family, or the volume. (This also retires an
earlier ``reverse-HyDE'' patch whose apparent $+0.023$ gain was a leakage artifact.)

\paragraph{Cross-provenance control: asymmetric transfer across generated-query distributions.}
Because the held-out queries are themselves SLS variants, the matched-budget gap could reflect
index--test provenance match (a recognized inflation source \citep{lewis2021testtrain}), not coverage.
We probe this with a \emph{cross-provenance} control on \textsc{Auto} (answer-in-index). After holding
out doc2query variants for testing, the resulting index sizes differ ($16.4$ doc2query vs.\ $23.5$ SLS
variants/unit), so this comparison does not remove index-volume as a possible contributor:
on doc2query's own held-out queries the SLS index reaches $0.652$, within $0.146$ of doc2query's own
index ($0.798$); but on SLS's held-out queries the doc2query index reaches only $0.531$, $0.399$ below
SLS's $0.930$ (hybrid \texttt{R@1}; the asymmetry is robust across BM25/dense). The
home-field advantage is thus strongly \emph{asymmetric}: SLS transfers to doc2query's distribution
better than the reverse. This is consistent with broader phrasing coverage, but the unequal index sizes
mean the control is suggestive rather than a causal isolation of provenance or volume.

\paragraph{Offline expansion adds no serving cost.}
Because the variants are attached at index time, serving is a single retrieval pass: on one NVIDIA H100
(Intel Xeon 8480C for BM25), median retrieval latency is $0.7$/$13$/$14$\,ms for BM25/dense BGE-M3/hybrid, whereas
query-time HyDE/query2doc add a \texttt{gpt-4.1-mini} call at $1.0$--$1.4$\,s, two orders of magnitude
more.

\begin{table}[t]
\centering
\footnotesize
\setlength{\tabcolsep}{4pt}
\resizebox{\columnwidth}{!}{%
\begin{tabular}{lllcc}
\toprule
Augmentation & side & Query-time generation? & Elec & Auto \\
\midrule
none (RAW)               & ---     & no  & 0.609 & 0.588 \\
doc2query ($6$-shot)     & offline & no  & 0.647 & 0.610 \\
doc2query ($N$-matched)  & offline & no  & 0.685 & 0.611 \\
EnrichIndex              & offline & no  & 0.662 & 0.616 \\
HyDE                     & online  & yes & 0.650 & 0.576 \\
query2doc                & online  & yes & 0.663 & 0.639 \\
\textbf{SLS (ours)}      & offline & \textbf{no}  & \textbf{0.881} & \textbf{0.930} \\
\bottomrule
\end{tabular}}
\caption{Offline-vs-online augmentation comparison, hybrid \texttt{R@1} on the same held-out query
variants ($319$ units). All methods, including SLS, use the \emph{same} \texttt{gpt-4.1-mini}.
Automatic expansions lift RAW only modestly; the matched-budget control (``doc2query
$N$-matched''---the \emph{same} generator producing the \emph{same} number of questions per unit as
SLS) reaches only $0.685$/$0.611$, at most $+0.038$ above the $6$-shot version. SLS (staged linguistic seeding)
dominates by $+0.20$/$+0.29$ over the best automatic baseline (online or offline), at zero
query-time generation cost. With the generator \emph{and} the per-unit budget held fixed, the gap is
attributable to the seeded recipe, not the model or the volume of generated text.}

\label{tab:bakeoff}
\end{table}

\paragraph{External validity.}
Because our data is proprietary, we replicate both protocol-level findings on the public Quora Question
Pairs set \citep{quora_qqp}: the leakage pitfall reappears and the expansion gain replicates on
lexical/hybrid retrieval but dilutes on a near-ceiling dense baseline (numbers in App.~\ref{app:qqp}).

\paragraph{Routing: a bounded secondary result.}
\label{sec:routing}
Weak-evidence queries should be routed (clarify/abstain/handoff) rather than answered. A class-balanced
gradient-boosted classifier on non-leaky retrieval-confidence features lifts action macro-F1 from $0.36$
to \textbf{0.51} ($0.45$ using retrieval features only, avoiding circularity) and recovers the rare
abstain/handoff actions; but the deployable signal is the \emph{binary} selective-answering slice, the
four-way result is bounded by heuristic, genuinely subjective labels ($\kappa{\le}0.09$ on
re-annotation by a question-only \texttt{gpt-5} and a human) and by features that indicate \emph{whether}
but not \emph{why} to escalate. We thus report routing as a bounded secondary study; full numbers, the
cost--coverage curve, and Table~\ref{tab:routing} are in App.~\ref{app:routing}.

\paragraph{Does verified-unit answering reduce hallucination?}
\label{sec:faithfulness}
We compare two answering arms on $150$ answerable queries given the \emph{same} top-5 retrieved units:
\emph{free-form RAG} (a generator writes the answer) vs.\ \emph{verified-unit answering} (return the
top-1 unit's answer verbatim), judged by an independent \texttt{gpt-4o} for unsupported claims and gold
coverage. On the correct-retrieval subset ($n{=}136$), free-form introduces unsupported content in
$7.4\%$ (\texttt{4o-mini})--$13.2\%$ (\texttt{4.1-mini}) of answers, whereas verified-unit answering is
unsupported-free \emph{by construction} (a residual $0.7\%$ is one judge-noise flag on verbatim text;
Table~\ref{tab:faithfulness}). This reduction is significant (paired McNemar $p{=}0.012$,
$7.6{\times}10^{-5}$), and because gold coverage is comparable ($81.6\%$ vs.\ $74.3\%$/$80.9\%$), the
gain comes from eliminating the \emph{unsupported-generation surface} rather than from coverage. As noted, the judge is LLM-based with no human adjudication (\S\ref{sec:limitations}).

\begin{table}[t]
\centering
\footnotesize
\setlength{\tabcolsep}{5pt}
\resizebox{\columnwidth}{!}{%
\begin{tabular}{lcc}
\toprule
Answering arm (same correct evidence, $n{=}136$) & Unsupported & Covers gold \\
\midrule
Free-form RAG (\texttt{gpt-4o-mini})  & 7.4\% & 74.3\% \\
Free-form RAG (\texttt{gpt-4.1-mini}) & 13.2\% & 80.9\% \\
Verified-unit (ours)                  & \textbf{0.7\%}$^{\dagger}$ & \textbf{81.6\%} \\
\bottomrule
\end{tabular}}
\caption{Faithfulness on the correct-retrieval subset ($n{=}136$): two free-form generators vs.\
verified-unit answering over the \emph{same} top-5 evidence (a single \texttt{gpt-4o} judge scoring unsupported-content and gold-coverage).
Verified-unit is unsupported-free \emph{by construction} ($^{\dagger}0.7\%$ is one residual judge-noise
flag on verbatim text). Paired McNemar on unsupported: $p{=}0.012$/$7.6{\times}10^{-5}$.}
\label{tab:faithfulness}
\end{table}

\section{Discussion}
\label{sec:discussion}

\paragraph{Why human-authored expansion outperforms learned query generation.}
A learned query generator (doc2query) or a thesaurus operates over \emph{surface form}: the
distributional co-occurrence captured by word statistics \citep{firth1957}. But many real reformulations
are not lexical variants: the \textsc{Auto} query ``how do I set my GPS to get there?'' requests a
dealership's \emph{address} yet shares no content word with ``where is the showroom?''
(Figure~\ref{fig:overview}). The link is world knowledge (a navigation destination \emph{is} an
address), so recovering it means passing through the referent, which a lexical resource cannot do. This
is a gap between surface form and meaning (an instance of symbol grounding, \citealp{harnad1990}).

Our protocol supplies this grounding by construction (Figure~\ref{fig:overview}): an operator factorizes
the representative question into orthogonal world-grounded slots (explicit for \textsc{Auto},
App.~\ref{app:example}; for \textsc{Elec} only the deployed variants survive), and the language model
\emph{recombines} these grounded units, sampling \emph{within} the operator-defined manifold rather than
guessing it, unlike free-form generation, which crowds near the seed or drifts off-target.

This account predicts \emph{where} the gain should land. On \textsc{Auto}, $49\%$ of
held-out queries share \emph{zero} content words with their unit's representative question, so a surface
match to that canonical form cannot reach them (BM25 \texttt{R@1} $0.28$ on this disjoint half). Indexing
the staged-seeding variants lifts exactly this half from $0.28$ to $0.87$ ($+0.59$); the gain then falls
monotonically with surface overlap: $+0.34$ for partially-overlapping queries ($n{=}378$) and $+0.01$
for the fully-overlapping ones already near ceiling ($n{=}91$). The gain thus concentrates on the
surface-disjoint, world-knowledge-dependent reformulations the account predicts, and the matched-budget
and matched-budget control (\S\ref{sec:results}) show that automatic expansion at the same model and
budget does not recover them; the cross-provenance result provides complementary, but index-size-confounded,
evidence of asymmetric transfer. We are careful about \emph{causality}: this shows a \emph{lexical resource}
cannot reproduce the expansion, not that a human beats an LLM \emph{generator} (which also has world
knowledge), the active ingredient is the world-grounded recipe, not the language-model call that renders
it.

\section{Limitations}
\label{sec:limitations}

Our faithfulness judge (\S\ref{sec:faithfulness}) is LLM-based with no human adjudication, so the
absolute unsupported rate is judge- and generator-dependent ($7$--$17\%$); the \emph{relative} reduction,
significant on correct-retrieval cases, is the robust finding. The retrieval study covers only 319
verified units (\textsc{Auto} just 90), so we report per-query and cluster bootstrap CIs. Our retrieval/faithfulness study is offline, without controlled A/B, and latency is a single-query micro-benchmark. The routing
result rests on \emph{heuristic}, subjective labels ($\kappa{\le}0.09$ on re-annotation) and features
that bound per-action scores, so we present the four-way router as forward-looking and report only the
binary answer-vs-escalate slice as deployable. We have not yet run a gate-only ablation (steps
(i)--(ii) without the step-(iii) gate) to separate the human gate from the seeding design, nor quantified
human authoring effort; the recipe is a \emph{one-off offline} asset that should amortize against the
recurring cost of a wrong answer, but a precise cost--benefit accounting and scaling to far larger FAQ
sets remain future work. The \textsc{Elec} recipe was not retained, so recipe-level analysis
(\S\ref{sec:discussion}) is \textsc{Auto}-only, though expansion was applied to both domains and gains
hold across both; raw logs remain access-controlled. Most fundamentally, the world-grounded recipe is
presently \emph{human}-authored; automatically \emph{inducing} this factorization is the central open
problem, and until then operator authoring is what bridges the curated-question/real-user gap.

\section{Ethics}

The data come from enterprise logs; we report only aggregated statistics, anonymized examples, and
unit IDs, and keep raw logs access-controlled. The method is not a truth guarantee (a wrong
retrieved unit can still produce a wrong response), so deployments should log unit IDs, confidence,
and action decisions for audit.

\section{Conclusion}

We presented a verified-unit QA pipeline for AI contact centers: answers are bounded to verified
units; weak-evidence cases are routed, not fabricated. Its lever is \emph{staged
linguistic seeding} that lifts hybrid \texttt{R@1} to
$0.881$/$0.930$ ($+0.27$/$+0.34$) across two domains; the matched-budget control attributes the main
gain to the seeding \emph{design}, while cross-provenance evaluation shows suggestive asymmetric transfer;
verified-unit answering removes the unsupported-generation surface
($\approx\!0$ vs.\ $7$--$13\%$). Inducing the recipe automatically remains open.

\section*{Acknowledgments}

We thank Dong-su Kim and Won-Chang Shin for their personal encouragement and operational feedback.
The authors received no dedicated financial support for this work.

\bibliography{groundlm_refs}

\clearpage
\appendix

\section{The Question-Identity Leakage Audit}
\label{app:leakage}
An earlier indexing protocol of ours was leaky. Indexing each unit as
``\texttt{question:}\,$\langle$representative question$\rangle$; \texttt{answer:}\,$\langle$answer$\rangle$''
(the question and answer fields are Korean in the deployment) and then querying with that \emph{same}
representative question makes retrieval match a query to its own indexed copy. On the full set of
$6{,}000$ clustered operational-log queries, the leaky question-bearing index scores BM25/dense/hybrid \texttt{R@1}
$0.863$/$0.570$/$0.873$, but only $0.103$/$0.222$/$0.160$ once the question field is removed, an
inflation of roughly $8\times$ for BM25 and $2.6$--$5.5\times$ across families
(Table~\ref{tab:leakage}). Closed-set FAQ evaluations in which test queries coincide with indexed
representative questions should be read with this inflation in mind; our held-out-variant protocol
avoids it because the test phrasing is never in the index.

\begin{table}[t]
\centering
\footnotesize
\setlength{\tabcolsep}{5pt}
\resizebox{\columnwidth}{!}{%
\begin{tabular}{lccc}
\toprule
R@1 ($6{,}000$ log queries) & BM25 & Dense & Hybrid \\
\midrule
Leaky (question-bearing index) & 0.863 & 0.570 & 0.873 \\
Honest (answer-only index)     & \textbf{0.103} & \textbf{0.222} & \textbf{0.160} \\
\bottomrule
\end{tabular}}
\caption{Answer-only leakage audit on one consistent run over the $6{,}000$ clustered operational-log queries (distinct from the 319 verified units). Indexing the representative
question and querying with it (``leaky'') vs.\ indexing the answer only (``honest''). Inflation is up
to $\approx$8$\times$ for BM25 and $2.6$--$5.5\times$ across families; the headline result
(Table~\ref{tab:main}) instead uses held-out query variants, which never appear in the index.}
\label{tab:leakage}
\end{table}

\section{Routing: Full Results}
\label{app:routing}
We treat routing as a cost-sensitive decision over the non-leaky evidence vector $z$ (per-family top and
top-2 scores, margins, ratios, cross-family top-unit agreement). A class-balanced gradient-boosted
classifier reaches macro-F1 \textbf{0.51} (5-fold $0.487{\pm}0.027$) and recovers the rare actions
(abstain/handoff F1 $0.41$/$0.39$, up from $0.13$/$0.17$ for a Gaussian naive-Bayes baseline on the same
features; Table~\ref{tab:routing}); using retrieval features only (to avoid circularity with the
surface cues that define the heuristic labels), it still reaches \textbf{0.45}.

\emph{Cost--coverage.} Under an asymmetric cost matrix (we adopt the \emph{illustrative} ratio that a
wrong auto-answer costs $20\times$ a clarification) and honest retrieval (answer-only
\texttt{hit@1}$\approx$0.13), the
expected-cost-minimizing policy is conservative: it auto-answers $<\!1\%$ of queries at a $0\%$
unsafe-answer rate (mean cost $1.10$ vs.\ $16.5$ for always-answer). Ranking answerable queries by
predicted retrieval-correctness, the top $10\%$ are answerable at precision $0.65$ (vs.\ a $0.13$ base
rate), falling to $0.34$ at $30\%$ and $0.24$ at $50\%$ coverage.

\emph{Two ceilings.} (i) The action labels are heuristic, not adjudicated: on a $300$-query sample, a
question-only \texttt{gpt-5} and a human re-annotation barely agree with them ($\kappa{=}0.06$/$0.04$)
or with each other ($\kappa{=}0.09$), and skew in opposite directions, so the four-way boundary is
genuinely subjective. (ii) Retrieval-confidence features separate \textsc{answer} from escalate but
carry little signal for \emph{why} to escalate, bounding four-way F1 independent of label quality. The
defensible result is therefore the binary selective-answering slice; the four-way router, with
human-adjudicated labels and intent-bearing features, is left to future work.

\begin{table}[t]
\centering
\footnotesize
\setlength{\tabcolsep}{4pt}
\resizebox{\columnwidth}{!}{%
\begin{tabular}{lccccc}
\toprule
F1 by action & answer & clarify & abstain & handoff & macro-F1 \\
\midrule
Gaussian NB (same features)  & 0.61 & 0.51 & 0.13 & 0.17 & 0.36 \\
Balanced GBM, retrieval-only & 0.62 & 0.54 & 0.28 & 0.38 & 0.45 \\
Balanced GBM, all features   & \textbf{0.66} & \textbf{0.58} & \textbf{0.41} & \textbf{0.39} & \textbf{0.51} \\
\bottomrule
\end{tabular}}
\caption{Action routing on the operational log (test $n{=}1{,}200$; non-leaky multi-family retrieval
features). Class-balanced gradient boosting lifts macro-F1 from $0.36$ (Gaussian NB on the same
features) to $0.51$ (all features) / $0.45$ (retrieval-only, the circularity-free number) and
recovers the rare \textsc{abstain}/\textsc{handoff} actions.}
\label{tab:routing}
\end{table}

\begin{table}[t]
\centering
\small
\setlength{\tabcolsep}{4pt}
\resizebox{\columnwidth}{!}{%
\begin{tabular}{lrrrr}
\toprule
Action label & Auto & Elec & Total & Share \\
\midrule
answer & 1,533 & 922 & 2,455 & 40.9\% \\
\midrule
\textbf{non-answer subtotal} & \textbf{2,067} & \textbf{1,478} & \textbf{3,545} & \textbf{59.1\%} \\
\quad clarify & 1,497 & 857 & 2,354 & 39.2\% \\
\quad handoff & 446 & 578 & 1,024 & 17.1\% \\
\quad abstain & 124 & 43 & 167 & 2.8\% \\
\midrule
\textbf{Total human queries} & \textbf{3,600} & \textbf{2,400} & \textbf{6,000} & \textbf{100.0\%} \\
\bottomrule
\end{tabular}}
\caption{Operational routing log with 6,000 human queries across two domains. Labels are heuristic
routing labels derived from rule-based buckets over answer content, not human-adjudicated gold labels.
Non-answer actions account for 59.1\% of the log, and abstain is rare (2.8\%), which makes rare-action
recovery central to the routing analysis (\S\ref{sec:routing}).}
\label{tab:data}
\end{table}

\section{External Validity: Quora Question Pairs}
\label{app:qqp}
Because our enterprise data is proprietary, we replicate both protocol-level findings on Quora Question
Pairs \citep{quora_qqp} ($800$ clusters as ``units'', $1{,}470$ held-out paraphrases, with each
cluster's other phrasings serving as the expansion). The leakage pitfall reappears: an indexed phrasing
gives \texttt{R@1} $1.000$ versus $0.81$--$0.94$ on held-out paraphrases, and the expansion effect
carries over to lexical and hybrid retrieval (BM25 $+0.124$, hybrid $+0.053$; cluster-bootstrap
significant), while a near-ceiling dense retriever on this clean set dilutes slightly ($-0.026$).
Expansion helps most where single-representative matching is weakest, exactly as on our enterprise logs.
This QQP study is the fully public, reproducible portion of our work; we will release its replication
code and derived data, together with our pipeline code and the fine-tuned checkpoint, upon acceptance. The
release pins the exact dated snapshot identifier and decoding parameters for every OpenAI call
(\texttt{gpt-4.1-mini} for SLS rendering and the automatic baselines; \texttt{gpt-4o} and \texttt{gpt-4o-mini}
for judging), so the SLS variants and judge labels regenerate deterministically.

\section{Dataset Statistics}
\label{app:data}

Table~\ref{tab:datastats} reports per-domain statistics for asset~(a), the verified units and their
indexed query-expansion variants. Lengths are measured on the original Korean text (characters; and
\emph{eojeols}, the space-delimited tokens of Korean).

\paragraph{Production deployment.}
The system is in live production on both domains: \textsc{Auto} since early 2025 (${\sim}18$ months) and
\textsc{Elec} since mid 2025 (${\sim}12$ months). Over these windows it has served $24{,}126$ \textsc{Auto}
requests (excluding reservation-management transactions) and $50{,}172$ \textsc{Elec} requests, routing
$3{,}423$ and $4{,}724$ respectively to a human agent: a containment (no-handoff) rate of $85.8\%$ and
$90.6\%$ ($89.0\%$ combined, over $74{,}298$ requests). Containment is a deflection measure, not a
correctness guarantee (non-handoff may include user abandonment), and we report no controlled before/after.

\begin{table}[ht]
\centering
\footnotesize
\setlength{\tabcolsep}{4pt}
\resizebox{\columnwidth}{!}{%
\begin{tabular}{lrr}
\toprule
 & \textsc{Auto} & \textsc{Elec} \\
\midrule
Verified units                       & 90    & 229 \\
Held-out test queries                & 915   & 1{,}468 \\
Indexed variants                     & 2{,}114 & 3{,}450 \\
Indexed variants / unit (mean; min--max) & 23.5 (6--79) & 15.1 (3--41) \\
Answer length (chars; mean, max)     & 80, 193 & 137, 903 \\
Rep.\ question length (chars; mean)  & 19    & 21 \\
Variant length (eojeols; mean)       & 5.1   & 5.5 \\
\bottomrule
\end{tabular}}
\caption{Per-domain statistics of the verified-unit corpus (asset~(a)). \textsc{Auto} has more
variants per unit and shorter, more templated answers; \textsc{Elec} has more units and much longer
answers (up to $903$ characters).}
\label{tab:datastats}
\end{table}

\section{An Example Verified Unit}
\label{app:example}

The example below is from \textsc{Auto}, anonymized and translated from Korean. It traces one
representative question through staged linguistic seeding: the \emph{seeding axes} that step~(i)
applies, a sample of the step~(ii) generations that the step~(iii) acceptance gate kept in the index,
and the held-out test queries (never indexed) the system is evaluated on.

\begin{quote}
\footnotesize
\textbf{Representative question:} ``What region are you in?''\\
\textbf{Verified answer:} ``$\langle$dealership$\rangle$ is located at $\langle$address$\rangle$.''
(anonymized)\\[2pt]
\textbf{Seeding axes (step i):} location-noun synonyms (region\,/\,location\,/\,address\,/\,lot-number),
inflectional and honorific endings, and spacing (eojeol) segmentation of the representative
question---the fixed seed specification handed to the generator.\\[2pt]
\textbf{Accepted variants (steps ii--iii, sample):} ``Where are you located?''; ``Can you tell me the address?'';
``What is the lot number?''; ``How can I find the location?''---each realizing one or more of the
step-(i) seed axes in the original Korean.\\[2pt]
\textbf{Held-out test queries (never indexed):} ``Please tell me the location.''; ``What is the
address of the showroom?''; ``Where is it located?''
\end{quote}

\section{How the Data Explains the Results}
\label{app:interp}

The statistics in Appendix~\ref{app:data} and the example in Appendix~\ref{app:example} account for the
main retrieval findings. First, a representative question is short (mean $19$--$21$ characters) and
fixes a single surface form, whereas held-out queries restate the same intent with different content
words and inflections; single-representative matching (RAW) therefore misses most of them, and the
$15$--$24$ indexed variants per unit close exactly this gap. This is why AUG lifts \texttt{R@1} so
sharply, and why the purely lexical BM25 gains the most (it keys directly on the added word forms).
Second, \textsc{Auto} reaches a higher AUG \texttt{R@1} ($0.930$) than \textsc{Elec} ($0.881$): it has
more variants per unit ($23.5$ vs.\ $15.1$), fewer units ($90$ vs.\ $229$), and shorter, more
templated answers (mean $80$ vs.\ $137$ characters), so its units are more densely covered and more
separable. Third, \textsc{Elec} answers are much longer (mean $137$, max $903$ characters), giving a
free-form generator more room to add unsupported detail; verbatim verified-unit answering returns the
exact text and stays unsupported-free, consistent with the faithfulness result (\S\ref{sec:faithfulness}).

\section{Domain-Adapted Checkpoint: Training Details and Korean Competence}
\label{app:ckpt}

The ``Dense (fine-tuned, ours)'' row in Table~\ref{tab:main} is our own checkpoint, reported as a
negative result. We deliberately did \emph{not} start from the vanilla official model: the base is
\texttt{dragonkue/BGE-m3-ko} \citep{dragonkuebgem3ko}, the official \texttt{BAAI/bge-m3}
continued-trained on additional, undisclosed Korean data. This base is \emph{Korean-specialized} and is documented to
improve over the official model on Korean retrieval---its model card reports $+0.09$ nDCG on a Korean
(financial) AutoRAG benchmark over the official checkpoint (with a smaller $-0.02$ on
MIRACL-Wikipedia, i.e.\ a domain-specific Korean gain). On top of it we fine-tune with the
sentence-transformers \texttt{MultipleNegativesRankingLoss} (in-batch-negative contrastive) on
(\texttt{query}, \texttt{positive-document}) pairs built from \textsc{Nemotron-Personas-Korea}
\citep{nemotronpersonaskorea}: $4{\times}$H100, $3$ epochs, per-device batch size $8$ with gradient
accumulation $4$, learning rate $2{\times}10^{-5}$, warmup ratio $0.1$, weight decay $0.01$,
\texttt{bf16}. The resulting checkpoint is itself a competent Korean retriever (AutoRAG
\texttt{MRR}\,$0.78$, \texttt{Hit@10}\,$0.95$; MIRACL \texttt{mRR}\,$0.58$), so its underperformance on
our closed-unit customer-service units (Table~\ref{tab:main}) is a genuine domain-transfer failure,
not a weak-model artifact: starting from a strong Korean base and adding persona fine-tuning still does
not beat the off-the-shelf official BGE-M3 with offline SLS expansion. The checkpoint will
be released upon acceptance.

\section{Baseline Generation Prompts}
\label{app:baselineprompts}

Every automatic expansion baseline in Table~\ref{tab:bakeoff} uses the \emph{same}
\texttt{gpt-4.1-mini} as SLS, is written natively in Korean, and is given contact-center framing
matched to the task; the Korean originals ship with our code release. English glosses:
\begin{itemize}\itemsep2pt
\item \textbf{doc2query} (index-side, from the answer): ``Given a contact-center FAQ answer, generate
  six Korean questions a real customer might ask to obtain this answer; output JSON only.'' Conditioned
  on the unit's answer (first $1{,}200$ characters). ``$N$-matched'' uses the identical prompt with the
  per-unit count raised to match SLS.
\item \textbf{EnrichIndex} (index-side, enriched passage): ``Expand the given FAQ answer so it is
  easier to retrieve: include key terms, synonyms, expressions users might use, and related subtopics,
  in one $2$--$4$ sentence paragraph; do not invent new facts---ground it in the answer.''
\item \textbf{HyDE} (query-side, hypothetical answer): ``For the following customer question, write a
  hypothetical short Korean answer ($1$--$2$ sentences) a contact center might give.''
\item \textbf{query2doc} (query-side, expanded query): ``Expand the following customer question to aid
  retrieval by appending related expressions and key terms of the same intent, in $2$--$3$ sentences.''
\end{itemize}
Because every baseline is Korean-native and contact-center-framed under the same generator, the SLS
advantage in Table~\ref{tab:bakeoff} is not an artifact of prompt language or domain framing.

\end{document}